Subramaniam Vincent[1], Jingsen Wang[2], Zhan Shi[2], Sahas Koka[3], Yi Fang[2]

[1] Markkula Center for Applied Ethics, Santa Clara University, Santa Clara, CA

[2] Department of Computer Science and Engineering, Santa Clara University, Santa Clara, CA

[3] Dublin High School, Dublin, CA

Correspondence concerning this article should be addressed to Subramaniam (Subbu) Vincent, Markkula Center for Applied Ethics, Santa Clara University, 500 El Camino Real, Santa Clara, CA 95053, United States. Email: svincent@scu.edu.

**About the authors:** Subramaniam Vincent is Director of Journalism and Media Ethics at the Markkula Center for Applied Ethics, Santa Clara University. Phoebe Wang was a MS student at Santa Clara University when conducting this work. Zhan Shi is currently a Ph.D. student at the Department of Computer Science and Engineering at Santa Clara University. Sahas Koka is a senior at Dublin High School, Dublin, CA. Yi Fang is Associate Professor of Computer Science and Engineering at Santa Clara University.

Measuring Large Language Models Capacity to Annotate Journalistic Sourcing

**Abstract**

Since the launch of ChatGPT in late 2022, the capacities of Large Language Models and their evaluation have been in constant discussion and evaluation both in academic research and in the industry. Scenarios and benchmarks have been developed in several areas such as law, medicine and math (Bommasani et al., 2023) and there is continuous evaluation of model variants. One area that has not received sufficient scenario development attention is journalism, and in particular journalistic sourcing and ethics. Sourcing is a crucial pillar to all original journalistic output. Evaluating the capacities of LLMs to annotate stories for the different signals of sourcing and how reporters justify them is a crucial scenario that warrants a benchmark approach. In this paper, we propose a scenario, method and proof-of-concept to evaluate LLM performance on identifying and annotating sourcing in news stories on a five-category schema. We offer the use case, our dataset[1] and metrics and as the first step towards systematic benchmarking. Our accuracy findings indicate LLM-based approaches have more catching up to do in identifying all the sourced statements in a story, and equally, in identifying the varied types of sources. We find that spotting source justifications is an even harder task.

---

[1] Dataset: https://huggingface.co/datasets/subbuvincent/llms-journ-sourcing

# Measuring Large Language Models Capacity to Annotate Journalistic Sourcing

## Introduction

Since the launch of ChatGPT in late 2022, the capacities of general-purpose Large Language Models and their evaluation have been in constant discussion, both in academic research and in the industry. Scenarios and benchmarks have been developed in several areas such as law, medicine and math (Bommasani et al., 2023) for instance, with the Holistic Evaluation of Language Models (HELM) initiative, to perform continuous evaluation of model variants. There is both excitement and worry about the impact of Artificial Intelligence (AI) in the news media and the ongoing interest in newsroom-situated experimentation with LLMs and chatbots. Despite this, one area that has not received sufficient scenario development attention for benchmarking is real-world journalism, and in particular journalistic sourcing and ethics. A scenario represents a use case and consists of a dataset of instances, according to HELM.

The authority of journalism is founded on a robust connection between news and truth (Steensen et al, 2022). Journalism is a crucial truth-determination function in democracy (Vincent, 2023). Sourcing is a crucial pillar to original journalistic output. Without people, organizations, footage, documents and data serving as sources, journalists would not be able to do their truth-telling work. And the manner in which journalists attribute statements, claims, findings, conclusions, and broadly the content in their stories to sources and justify their inclusion is often unstructured and embedded in the communication itself, be it writing, audio or video. But assessing, even if roughly, whether and to what degree a series of stories is sourced from the democratic stakeholdership on issues requires new datasets based on source retrieval and annotations around an ethics vocabulary. Source retrieval, text extraction and analysis are in use for development of methodologies to study and visualize bias in source selection (Zhukova et al., 2023). With source and source-related definitions, categorization, and enumeration, the work of building quantitative measures and qualitative assessments of sourcing, for e.g. in news coverage of chronic issues such as homelessness (Moorehead, 2023), using standards from ethics is possible. Manual efforts are both expensive and difficult to instrument in real time when news cycles are ongoing. In the real world, this impoverishes the discourse on assessing the ethical quality of news stories - especially in technological distribution systems such as news aggregators and social media - to the realm of factual accuracy, as opposed to expanding to account for stakeholder inclusion, perspective-centering, including the lived experiences of communities.

Evaluating the capacities of LLMs to use a supplied vocabulary about journalistic sourcing to identify and annotate stories, and offering an approach to benchmarking is the core work behind this paper. The intersection of language and journalism is the topic of research (Jaakkola, 2018). We define the most salient elements of sourcing - statements attributed to sources, types of sources, names of sources, titles, justifications and characterizations of sources. LLMs have substantive language capacities and it ought to be possible to evaluate how well they identify sourcing language and various elements as a narrow task. This is a crucial scenario that warrants a benchmark approach because if LLMs could get the job done, it may open the door to expand the accessibility and customizability of annotation and/or audit solutions for different genres and formats of newsroom output.

Second, we are able to see which LLMs perform better for different types of source annotations and compare them. If there is significant underperformance, it means that the claims

Measuring Large Language Models Capacity to Annotate Journalistic Sourcing

of LLMs passing a bar exam, scoring high on school math exams, etc., are not useful to journalism evaluation.

In this paper, we propose a scenario, method and proof-of-concept to evaluate LLM performance on a five-category schema partly inspired from journalism studies (Gans, 2004). It involves identifying and annotating sourced statements, sources, their types, and source characterizations and justifications. We offer the use case, our dataset[1], the LLM prompts, our ground truth data, and a set comparison metrics for each model as an initial framework towards systematic benchmarking of LLMs for accuracy in journalistic sourcing review.

Our accuracy findings indicate LLM-based approaches have more catching up to do in identifying all the sourced statements in a story, and equally in the task of spotting source justifications. Source justifications in particular, if accurately annotatable, could provide a signal and incentive to help distinguish between the more bottom-up ethical grades of journalism from the top-down (expert and authority-sourced) forms.

**Related Work**

There has been substantive interest in the use of language models for analysis applications in journalism environments. One study (Bhargava, 2024) tested the capacity of ChatGPT to audit sources at a local university-based news outlet. Another study (Li et al., 2024) probed GPT-4 for knowledge of journalistic tasks and compared it to an existing database of occupation descriptions. This analysis was situated in the context of developing agentic systems for journalism. Another study (Spangher et al., 2024) has studied large language models (LLMs) for a role in longer-form article generation itself as part of an effort to explain how journalists plan their sourcing, amongst other workflow tasks, before writing. The closest to our work is a study

Measuring Large Language Models Capacity to Annotate Journalistic Sourcing

on identifying information sources in news articles (Spangher et al, 2023). The authors tested a fine-tuned GPT3 for identifying sourced statements and source attribution. They defined the *compositionality* of sources in a news story as a prediction problem. LLMs benchmarking for news summarization is also a topic of interest (Zhang et al., 2024). Our goals were to develop a reproducible benchmark to evaluate multiple LLMs for journalistic sourcing ethics. Our interest in types of sources (for detection) include everyday people (non-experts) and those in positions of power and formal expertise (titles), along with detecting textual justifications. Another difference is our work could afford much larger token budgets to test general purpose LLMs, allowing use of detailed prompts and definitions on instruction-tuned LLMs directly.

In (Vincent et al., 2023), the lead author of this paper and co-authors documented an effort to use source-diversity proportions data from direct quotes annotations for over 800 news sites to conduct boundary analysis for journalistic behavior online. The computational system was Stanford CoreNLP-based, and augmented with a machine learning model (Shang et al., 2022) for quote extraction and attribution. However, direct quotes are only one form of attribution to sources. The limitations of the earlier NLP-based technology to annotate more complex types of journalistic source attributions and justifications hamper prospects of building datasets for more comprehensive sourcing analysis and audits. This is one of the reasons for us to propose the need to benchmark LLMs around a new schema. Our initial effort was through an MS thesis work (Wang, 2024) where we staged and tested a preliminary set of LLM prompts, followed by this work.

We know of no current work aimed at proposing a new use case and scenario, on the lines of HELM's approach to test and compare general purpose LLMs for annotating journalistic sourcing as a capacity.

Measuring Large Language Models Capacity to Annotate Journalistic Sourcing

## Hypothesis

As we noted earlier, LLMs have substantive language capacities and a variety of claims are made about them through benchmarks in other scenarios such as law, medicine, math, reasoning, and so forth. Discerning the language used by journalists to show their sourcing work in stories is a tough area for annotations by LLMs, and we would like to examine the following hypothesis.

**H1a:** LLMs will be able to accurately spot different types of statements attributed to sources, not only direct quotes. Previous generations of NLP technologies such as Stanford Core NLP (Muzny et al., 2021) have had quote attribution modules to detect direct quotes in story texts. But journalistic news often has indirect speech and attributions in statements to anonymous sources, documents and unnamed groups of people. We expect that LLMs ought to be able to detect these additional types of sources as well, and would like to evaluate the overall accuracy for the different models.

**H1b:** LLMs will be to identify the different types of sources journalists use in news stories, given a set of plain text definitions for named persons, named organizations, unnamed groups of people, documents and anonymous sources. We test how the models apply the definitions to catch all the different types of sources in a story with their sourced statements. Types of sources and sourced statements are related and we are interested in testing both H1a and H1b together.

**H2a:** LLMs will be able to identify source titles and justifications since these attributes are components of language used by journalists and go to the heart of their general-purpose capacities on language. Journalists usually signal to readers why a source is being attributed in the story using their title or expertise, their presence at a meeting or event, their lived experience or historical connection to the issue. By offering definitions for the terms source titles and source justifications, we evaluate whether the LLMs are able identify these attributes and if so, how accurately.

**H2b:** Specifically, we also hypothesize that LLMs will be able to spot anonymous sourcing language, and extract the journalist's justification to attribute statements to the source. Anonymous source justifications are normatively significant because they involve disclosure about the sensitivity in the status or role of the human source who engaged with the journalist for the story. Journalists' use of anonymous sources was once scorned and later gained more acceptance despite ethical concerns (Duffy, 2014). But a lack of consistent scrutiny on the use of anonymous sources creates a blind spot not only in our understanding of news sources, but of journalism more broadly (Carlson, 2020). A validated capacity in LLMs to spot both anonymous sourcing and the presence or absence of justifications could have positive implications for downstream ethics audit applications.

**H3:** We expect the leading "open source" models (Llama, from Meta) to also perform as well as (or nearly) closed source models (Open AI, Anthropic and Google), especially the 405 Billion parameter model. The comparative performance of open-source LLMs for text classification in political science contexts has received attention for performance benchmarks (Alizadeh et al., 2024) because of the fine-tuning possibilities in these models.

## Models evaluated

We selected the following models for testing:

Measuring Large Language Models Capacity to Annotate Journalistic Sourcing

1. Anthropic's Claude 3.5 Sonnet

2. OpenAI's ChatGPT-4o

3. Google's Gemini Pro 1.5

4. Meta's Llama 3.1 405B Instruct, and

5. Meta's Llama 3.1 Nemotron 70B Instruct.

6. DeepSeek R1

We used Llama 3.1's two variants and DeepSeek R1 to include three "open source" variants, and the other three as "closed source" models. At the point when we started out on this research, we did not have any hypothesis about which model would perform the best in spotting sources, attributions and justifications.

**Methodology**

**Story selection and sample size**

To build a proof-of-concept for benchmarking LLMs, we selected a small-scale corpus, similar to Brigham et al. (2024). We selected 34 articles from a variety of different news publishers with 557 sourced statements that we identified through our ground truthing process. See Table 1 for the list. The only common aspect of these news outlets is that they are engaged in journalistic work. All of them follow conventional sourcing and attributional practices. They undertake both original reporting and also publish interpretative commentary based on facts and factual claims. They range from local to national to issue based and BIPOC-led. (Black, Indigenous and People of Color). Our interest is only in benchmarking LLMs on spotting the journalistic sourcing manifest in the content of these articles.

Measuring Large Language Models Capacity to Annotate Journalistic Sourcing

| Site name | Type of site | Number of stories in sample (total=34) |
|---|---|---|
| Regional/Local News | | |
| Cal Matters | CA news | 2 |
| Documented NY | Immigration News | 1 |
| LAIst | Southern California Public Media | 2 |
| MLK50 | Tennessee regional | 3 |
| Oakland Side | Bay Area Local News | 3 |
| Sacramento Observer | CA regional news, Black-owned | 2 |
| SF Gazetteer | SF Local News | 2 |
| SF Standard | SF Local News | 2 |
| VT Digger | Vermont Local News | 2 |
| WHYY | Philly Public Media | 1 |
| Mercury News | Bay Area Local News | 1 |
| Central Virginian | Virginia regional | 1 |
| | | |
| Non-local/Mission/National/Wire | | |
| 19th News | Mission oriented, gender and politics | 1 |
| Capital B | Black-owned | 2 |
| Native News Online | Indigenous American | 1 |
| Associated Press | National/Syndicated News | 4 |
| New York Times | National/International | 1 |
| Reuters | National/International | 1 |
| Salon.com | National/International | 1 |
| Reason.com | News and Opinion | 1 |
| **Total number of articles** | | **34** |

Table 1: List of publishers and articles in our news corpus.

Measuring Large Language Models Capacity to Annotate Journalistic Sourcing

**Input Statistics from the Article Samples (Workload to LLMs)**

Figure 1 below shows the distribution of types of sources in 557 sourcing statements across the 34 news articles. We developed the full ground truth data for the sourcing in all the stories, and this distribution is drawn from the aggregate. See the next section for the definitions of the types of sources.

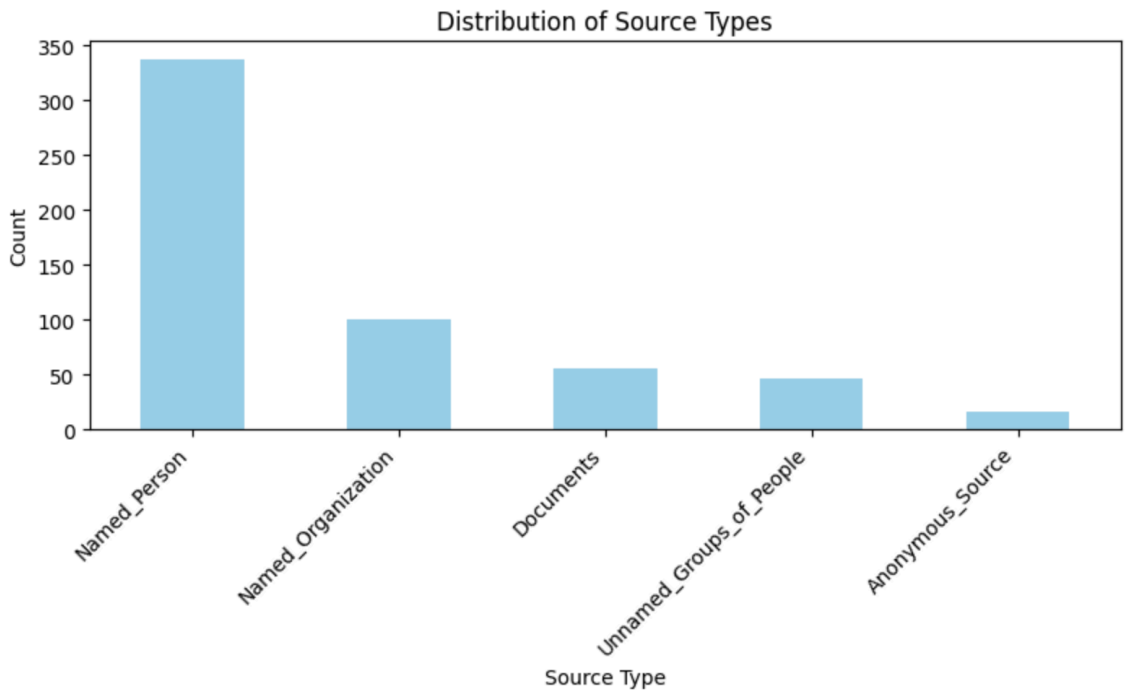

Figure 1. Distribution of types of sources in the workload used to test the LLMs.

**Our schema for journalistic source annotations**

We define the following five elements of journalistic sourcing as critical attributes to look for in news stories. 1) Sourced Statement 2) Type of Source 3) Name of Source 4) Title of Source (when present), and 5) Source Justification (when present).

Source titles and source justifications are key distinguishing aspects signaling the ethics of the sourcing process. Those sources are usually introduced with their title and additional qualifications in stories, interviews, podcast conversations and so forth. Democratic pressure on sourcing results in people and communities being included and centered in stories. Those sources may or may not have titles (expertise or power), but there is an expectation from ethics that their inclusion be justified (or accounted for) all the same.

In a story (Guha, 2024) about a new law passed in the Vermont state legislature banning hair discrimination, the reporter not only cited lawmakers by title and constituency, but also that some of them were co-sponsors of the bill. There were also sources quoted as people who experienced discrimination or as parents of children who did. These are examples of source justifications.

Through the system prompt, we provided all the foundational definitions for the following terms to the LLMs: Source, Types of Sources (Named Person, Named Organization, Document, Anonymous Source, and Unnamed Group of People). We used this to insert all the data definitions for the five-attribute annotation: Sourced Statement, Name of Source, Title of Source and Source Justification. We asked the LLMs to take on the role of an analyst. These definitions were revised repeatedly through an iterative process to the final versions to generate the data for the calculations. They are written in plain English, as if explaining sourcing terminology to a high school student. Here is an excerpt from one definition for the term "source".

**"Source:** A source in journalism is a person, organization, document, or another news article from whom a journalist takes viewpoints, experiences, claims, expertise, positions, insights, knowledge, data or documents. Reporters may directly quote their sources or use indirect speech to paraphrase a source's views or claims. Sometimes the source is the person who sent sensitive material such as emails or documents or other internal organizational correspondence to the reporter. The source is a person who may have been present at meetings where sensitive deliberations or discussions took place, and that person then shared material from the meetings with the journalist. For e.g., when a reporter attributes a claim or statement to a person or a group of people with words such as "according to people familiar with..", or "according to people who were present at the meeting..", it means that person or those people are the source. When the reporter attributes claims or a statement using words like "according to a copy of emails reviewed by this newspaper" or "according to a copy of the document.." it means the source could be a document. If the person who sent the emails or documents to the reporter is granted anonymity by the report, then that person is an anonymous source. (See also, our definition of Anonymous Source.)" (Full definition in the dataset's system prompt file[2].)

Another part of our schema that illustrates the goal of this project is the definition we offer in the system prompt for "source justification". Here it is verbatim.

**"Source Justification:** This refers to any additional source characterization or explanation that justifies to the reader why the source is in the story or that section of the story, how they are connected to the story and/or to other sources in the story. Any of the five defined types of sources may have such justifications and explanations present. It is not the same as the title of source, which is the previous definition above. It may be a few words, a part of a sentence, multiple sentences, or a full paragraph. The reporter will usually offer a justification in the story when they introduce the source. Source justification may be a part of the sourced statement itself. Sometimes the source justification comes later or earlier in the article where the

Measuring Large Language Models Capacity to Annotate Journalistic SourcingMeasuring Large Language Models Capacity to Annotate Journalistic Sourcing source or some situation involving the source is referred to. While source justification is NOT the same as title of the source, it may include the title of the source. For named persons, the source justification text may narrate the lived experience of the source. When sources are people who are stakeholders to the issue being reported on, who witnessed something happen, or have a lived experience related to the issue of the story, or a co-litigant in a lawsuit, etc., narrating this demonstrates their significance in the story for readers. For e.g., someone who went through a period of homelessness may be quoted for their lived experience and opinion about solutions. Someone else may have spent four years waiting to get a job or to get their voting rights back because of a prior felony conviction. Remember that Source Justification is not the same as title of the source, but may include both the title of the source and the explanations for the source's role in the story or relationship to other sources. Named Persons or Anonymous Sources without a title may still have source justification present in the text."

To meet length limits for the paper, we have not included the other definitions. The full list of is available in the system prompt[2] file in the dataset.

**Ground Truth development**

We took the following approach for developing ground truthing data for all for the 34 stories. We offered six volunteer graduate students a brief orientation and explanations of the same journalistic sourcing vocabulary (as given in the prompts for LLMs). They then annotated two stories as their first attempt, which we reviewed together. This process leads to corrections, removal of misunderstandings and final version of the sourcing annotations on the five-element

---

[2] https://huggingface.co/datasets/subbuvincent/llms-journ-sourcing/blob/main/prompts/system_prompt_v40.txt

Measuring Large Language Models Capacity to Annotate Journalistic Sourcing

schema. After this, the students annotated the production articles in our data sample. The lead author then reviewed all of the annotations sheets per story individually, made corrections and produced one ground truth data file per story.

Each ground truth data file (CSV format) has a list of sourcing statements (rows), with type of source, name of source, title, and source justification. Table 2 shows the layout for the ground truth annotations, one per news article.

| S.No | Sourced Statement | Type of Source | Anonymous? Y/N | Name of Source | Title | Source Justification |
|---|---|---|---|---|---|---|
| | | | | | | |

Table 2: Layout for the ground truth annotations, per news article.

The lead author originally tested this approach in an internship course at Santa Clara University in Spring'24 (COMM 198) with a group of undergraduate students from humanities and engineering. This led to both experiential learning on sourcing literacy amongst the students and the validation that ground truth data production for sourcing could be developed for LLM benchmarking efforts.

Table 3 below is a data clip from the ground truth data file containing sourcing annotations for one of the 34 stories. This story (article 31) reported on a law passed in the state of Vermont prohibiting race-based hair discrimination.

| Sourced Statements | Type of source | Anonymity Y/N | Name of Source | Title of Source | Source Justification |
|---|---|---|---|---|---|
| A senior at North Country Union High School in Newport, Wilburn, 17, recalls being in line for the bathroom when the girl in front turned around and reached for her hair. Despite telling her not to | Named | | Aaliyah Wilburn | leader with the Vermont Student Anti-Racism Network | senior at North Country Union High School |

Measuring Large Language Models Capacity to Annotate Journalistic Sourcing

| | | | | |
|---|---|---|---|---|
| touch it, Wilburn said the girl "grabbed" her hair. | | | | |
| She cited a 2023 study that found 66% of Black girls in predominantly white schools and 44% of Black girls in all schools report experiencing hair discrimination, and that the experiences typically happen before they are 10. | Named | | Saudia LaMont | Rep. D-Morristown | said during a preliminary vote on the bill |
| A teacher explained that her daughter was reacting to several instances of students touching and petting her hair without her consent, LaMont continued. | Named | | Saudia LaMont | Rep. D-Morristown | said during a preliminary vote on the bill, comment from daughter's preschool teacher |

Table 3: Partial example of ground-truth annotations for article 31 in our sample.

**The User Prompt: Logic and learnings**

This user prompt is where the five models asked to do the real work of parsing the story and producing the data in a series of steps, using the definitions given in the system prompt. The full user prompt in the dataset[3].

By reviewing what the LLMs were able to catch and what they missed or annotated inaccurately, we are able to revise the prompts over and over again. We stopped at version 40. Our learnings about the sourcing language parsing capacities of LLMs from the revisions are:

1. The more detailed the definitions, with examples, the more likely the LLM will be able to apply them comprehensively to the article. This is why our definitions are quite detailed, more like descriptions, with examples.

2. In our initial versions, we asked LLMs to apply all the type of source definitions to the

---

[3] https://huggingface.co/datasets/subbuvincent/llms-journ-sourcing/blob/main/prompts/user_prompt_v40.txt

article at the same time (one instruction), identify the corresponding sourced statements, and the rest of the data (name of source, title, justifications, etc.) This results in non-comprehensive annotations where many sourced statements are left out, and often unpredictably.

3. We redesigned the prompts to instruct the models to parse the article for one type of source after another, in serial order. This modularizes the instructions, allows us to test per type of source, and produces more comprehensive results. After each type of source pass, we ask the models to generate the JSON data element for that type of source. And so on, till we finish all types of sources.

4. We also discovered that when the sources that "easier" to identify for humans, for e.g. named persons or named organizations, come later in the sequence, the comprehensiveness of the sourced statements identified improves. For e.g. we found that anonymous sourcing is harder to pick up, even for humans without training, because of the inherently unstructured and non-obligatory nature of standardized disclosure about sources in journalistic writing. Refer to our definitions in the system prompt. For anonymous sourcing, we have given cues in the definition and examples that require the LLMs to look for attributions to people who are not being named, (without or without the language that they requested anonymity for acceptable reasons), and where those people were somehow crucial to letting the reporter access viewpoints, claims, developments, decisions, etc., that they might have been witness to, or have documentation about. We also included in the definition that anonymous refers only to the public status of the source who is unnamed in story to protect the identity of the source, but that the journalists and often editors know the person. Reporters often signal that using language such as "three people who were present at the meeting..", etc. Compare this to a different type of source, also involved in attributed statements, where reporters refer to groups of people at an event doing or saying something together. These are unnamed groups of people, a different type of source, where the people did not seek anonymity from the journalist as part of the sourcing engagement.

5. In addition, for anonymous sources, our initial prompts did not include examples about how reporters signal them. For e.g. when internal documents or footage from an organization is sent by a source to a journalist, they may contain a named person making a newsworthy claim. In response to our initial definitions, all five LLMs would pull the claim, but mistake the type of source as the named person in the source-sent material, as opposed to the person (anonymous source) who sent it. For this reason, it helped for the Document type of source search to happen after the anonymous sources and unnamed groups of people are identified along with their sourcing statements.

7. Our final prompt instructs the models to parse the article for one type of source after another, step-by-step, (Wei et al., 2022) in serial order. This modularizes the instructions, allows us to test per type of source, and produces more comprehensive results. After each type of source pass, we ask the models to generate the JSON data element for that type of source. And so on, till we finish all types of sources.

   a. Anonymous sources (most difficult)
   b. Unnamed Groups of People.
   c. Documents.
   d. Named Persons.
   e. Named Organizations.

Measuring Large Language Models Capacity to Annotate Journalistic Sourcing

**Creating the LLM annotated data annotations**

Our article pre-processing code uses the Trafilatura library to extract article content and process the texts. It gets the webpage content using trafilatura.fetch_url, extracts metadata such as the headline, subtitle, publication date, and publisher through trafilatura.extract_metadata, and retrieves the main text of the article with trafilatura.extract. The extracted information is formatted and saved to one output text file per news article.

We developed a python script to streamline generating outputs from various LLMs using the OpenRouter API and format them into JSON and CSV files for metric evaluations. It begins by loading the system and user prompts, combining them with content from the article text files. These inputs are sent to LLMs via API requests, with mechanisms for retries to ensure valid responses.

Extracted JSON-formatted outputs are cleaned and validated, then saved alongside converted CSV files. The code processes multiple articles iteratively, saving results in experiment-specific directories, enabling structured and reproducible evaluations for comparative analysis.

We include two example snippets of generated annotations to illustrate how the models interpret the prompts on journalistic sourcing.

Gemini 1.5-Pro's annotation of one sourcing statement from article 4, "SFO Labor Day travel crowds are so back as United warns of staffing woes".

```
{
    "Sourced Statement": ""I have TSA Pre, so that should make it a
    little easier," said Bay Area resident Katelin Tharp, who on Thursday
```

```
was booted from an overbooked flight to Boston, where she's beginning

her fifth year at Northeastern University.",

    "Name of Source": "Katelin Tharp",

    "Type of Source": "Named Person",

    "Title of Source": null,

    "Source Justification": "Bay Area resident; booted from an

overbooked flight to Boston; beginning her fifth year at Northeastern

University"

}
```

Claude 3.5-Sonnet's of the exact same sourcing statement from that story:

```
{

"Sourced Statement": "\"I have TSA Pre, so that should make it a little

easier,\" said Bay Area resident Katelin Tharp, who on Thursday was

booted from an overbooked flight to Boston, where she's beginning her

fifth year at Northeastern University.",

"Name of Source": "Katelin Tharp",

"Type of Source": "Named Person",

"Title of Source": null,

"Source Justification": "Bay Area resident; beginning her fifth

year at Northeastern University"

}
```

In this case, the accurate source justification is Gemini 1.5 Pro, because it captured the reporter's key inclusion that the source was booted from an overbooked flight.

Measuring Large Language Models Capacity to Annotate Journalistic Sourcing

Another snippet below shows a fragment from Llama 3.1-405Billion-parameters' annotation of article 32 on OpenAI CEO Sam Altman's firing which relied substantially on document and anonymous sourcing. This illustrates this model's source justification extraction for statements attributed to anonymous sources. See dataset[4] for the full annotation.

```
{
"Sourced Statement": "After vetting four candidates for one position,
the remaining directors couldn't agree on who should fill it, said the
two people familiar with the board's deliberations.",
"Name of Source": null,
"Type of Source": "Anonymous Source",
"Title of Source": null,
"Source Justification": "two people familiar with the board's
deliberations"
},
{
"Sourced Statement": "Hours after Mr. Altman was ousted, OpenAI
executives confronted the remaining board members during a video call,
according to three people who were on the call.",
"Name of Source": null,
"Type of Source": "Anonymous Source",
"Title of Source": null,
"Source Justification": "three people who were on the call"
}
```

---

[4] https://huggingface.co/datasets/subbuvincent/llms-journ-sourcing/tree/main/llm_generated_annotations

*Handling LLMs unpredictability: Five generated JSONs per article*

Most publicly available LLMs have a setting called temperature with a range from 0 to 2. 0 tells the model to be most deterministic (or predictable). That said, we noticed during testing that even with a temperature setting of zero, there were minor variations in the JSON data generated by the LLMs when the same article was sent back for annotations to the same model. The differences were in sourced statements being missed or included, and at other times source justifications texts would be correctly spotted or left out. To account for this, we attempted to generate five sourcing annotations (JSON outputs, subsequently converted to CSV files) for each of the 34 news articles. We scored each CSV separately and averaged the scores per model per article and then built the overall score per model for all 34 articles.

*Total generated data from five LLMs for 34 articles*

We aimed at generating 850 annotation-carrying data CSVs in all. This is based on the following calculation: 34 stories x 6 models x 5 iterations (annotated data versions) per model = 1020 JSONs (or converted to CSVs).

In reality we generated 996 CSVs, because for a few articles the models produced 1-2 valid annotations, instead of five.

**The dataset**

Our dataset[1] has four folders, described in Table 4 below.

| Data item | Description |
| --- | --- |
| news articles for sourcing annotations | The plain texts of the 34 stories (input sample) |
| ground_truth annotations | The five-attribute (category) sourcing annotation data table, one for each of the 34 |

Measuring Large Language Models Capacity to Annotate Journalistic Sourcing

|  | sample stories, developed by our annotation team. |
|---|---|
| prompts | The system and user prompts developed and fed to all the five models tested. |
| llm_generated_annotations | The five-category sourcing annotation data generated (five versions per story) by each for the five models |

Table 4: Brief description of the dataset.

**Comparison functions to score the LLMs for each annotated data element**

We use three different matching functions to compare LLMs generated data (sourced statements, type of source, name of source, title and source justifications) to ground truth data, as described below. Table 5 shows the ground truth data to LLMs data comparison method for accuracy scoring.

| Ground truth data to LLMs data comparisons : methods for accuracy scoring | | | |
|---|---|---|---|
| Journalistic sourcing annotation element | Type of data | Matching function | Similarity threshold |
| Sourced Statement | Unstructured text | Semantic match | 0.8 |
| Type of Source | Structured text | Exact match | n/a |
| Name of Source | Structured, but minor variations are possible | Fuzzy match | 0.8 |
| Title of Source | Unstructured, may contain partial title or title with organizational affiliation | Semantic match | 0.55 |
| Source Justification | Unstructured, may be part of sourced statement, or text from other paragraphs | Semantic match | 0.55 |

Table 5: Ground truth data to LLMs data comparison method for accuracy scoring

Measuring Large Language Models Capacity to Annotate Journalistic Sourcing

**Fuzzy match:** The fuzzy match (Mouselimis L, 2021) function leverages a text similarity algorithm based on the Levenshtein distance, which measures how many single-character edits (such as insertions, deletions, or substitutions) are needed to transform one string into another. This approach returns a similarity score ranging from 0 (completely different) to 100 (identical), indicating how closely two text strings match.

The process begins by cleaning the text from both the ground truth and model-generated output to remove any formatting inconsistencies. Afterward, the similarity score is calculated, reflecting how similar the two cleaned texts are. This method is particularly useful for tasks like name matching, where textual variations such as misspellings, spacing differences, or capitalization changes may occur.

This function is applied for name matching. (Name of source metrics).

The threshold for fuzzy match is set at 80 to balance accuracy and tolerance for minor textual variations, ensuring that relevant matches are captured while minimizing false positives.

**Semantic match:** The semantic match (Reimers et al., 2019) function uses advanced natural language processing techniques to evaluate the meaning of text rather than its exact wording. It relies on a pre-trained language model that transforms text into numerical representations called embeddings. These embeddings capture the semantic essence of the text.

To compare texts, embeddings are generated for both the ground truth and the model-generated output. The similarity between these embeddings is then measured using a mathematical metric called cosine similarity. This score indicates how closely the two texts align in meaning, with higher scores reflecting greater semantic similarity.

For improved accuracy, longer texts are split into sentences. Each sentence from the ground truth is compared with every sentence from the model output, and the highest similarity score is used. This method is applied in task sentence matching, title matching, and justification matching, where capturing the underlying meaning is crucial.

To enhance the clarity and precision of the semantic match process, specific thresholds are set for evaluating similarity scores. A threshold of 0.8 is used for sentence matching to ensure a high level of semantic alignment, which is crucial for tasks where precise meaning is essential. This high threshold helps in maintaining a stringent standard of similarity, ensuring that the matched sentences share a strong contextual and semantic resemblance.

For title matching and justification matching, a threshold of 0.55 is utilized. This value strikes a balance between being overly strict and too lenient, allowing for a reasonable degree of semantic correspondence while accommodating minor variations in language expression.

**Exact match:** Our exact match function performs a straightforward comparison by checking whether two text strings are identical. Since the source types come from a predefined set, strict equality is required for this comparison.

**Accuracy formulae for the five sourcing attributes in the schema**

$$Source\_Statement\_Match\_Rate = \frac{Sentence\_matched\_num}{GT\_sentence\_num}$$

$$Source\_Type\_Match\_Rate = \frac{Type\_matched\_num}{Sentence\_matched\_num}$$

$$Name\_Match\_Rate = \frac{Name\_matched\_num}{Sentence\_matched\_num}$$

$$Name\_matched\_num = \frac{Title\_matched\_num}{Sentence\_matched\_num}$$

$$Justification\_Match\_Rate = \frac{Justification\_matched\_num}{Sentence\_matched\_num}$$

Measuring Large Language Models Capacity to Annotate Journalistic Sourcing

**Determination of average accuracy scores**

To develop the performance metrics for each LLM, we do the following:

1. For a given story, we calculate the accuracy score per sourcing attribute (sourced statement to source justification) using the functions above for a model.

2. Then we average the scores out for all five CSV samples per story, per sourcing attribute. That produces the model's score for that story and attribute. (For two articles in the 34-sample set, we found that the models do not produce 5 valid outputs for the iterations on the same story, and hence we have fewer than five CSVs.)

3. We then average the model scores across the 34 articles to produce a score for each sourcing attribute. We use that to compare the models at the per attribute level and as a whole.

## Results

We reviewed the accuracy scores per model for each of the five attributes of sourcing first. Following that we reviewed the overall model scores for accuracy across all attributes taken together.

**Sourced statement accuracy**

This is a comprehensiveness measure shown in Figure 2. It measures the extent to which the sourced statements identified by the five LLMs matched those in the ground truth CSVs. A 100% score for this metric would mean that the model pulled all of the sourced statements, across all of the stories as found in the ground truth. We find that Gemini 1.5 Pro scored 76.6% accuracy, with DeepSeek R1 coming in at 69.4%..

Measuring Large Language Models Capacity to Annotate Journalistic Sourcing

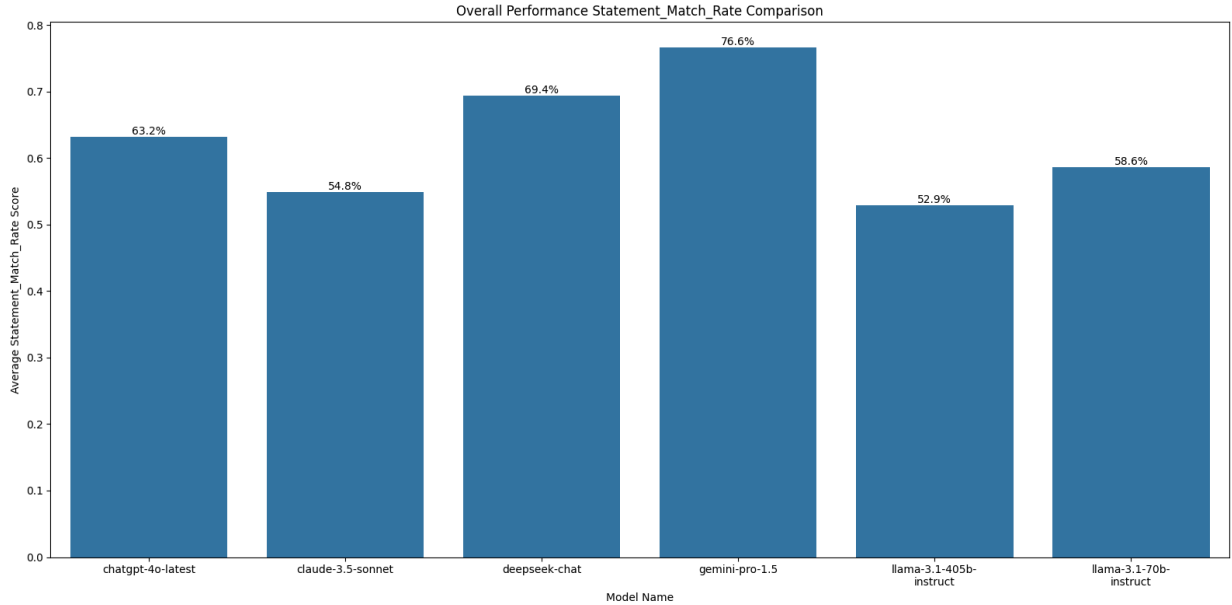

Figure 2: Sourced statement extraction accuracy comparison between the models.

**Type of source accuracy**

This is a key metric on LLMs identifying the type of source (named person, named organization, document, anonymous, or unnamed groups of people) accurately, as shown in Figure 3. Here we find that Claude 3.5 Sonnet, Gemini 1.5 Pro and Llama 3.1-405B all are in the 80-90% accuracy range, with Claude scoring 88.5%. GPT-4o, DeepSeek R1, and Llama 3.1-70B are in the 775-80% accuracy range.

Measuring Large Language Models Capacity to Annotate Journalistic Sourcing

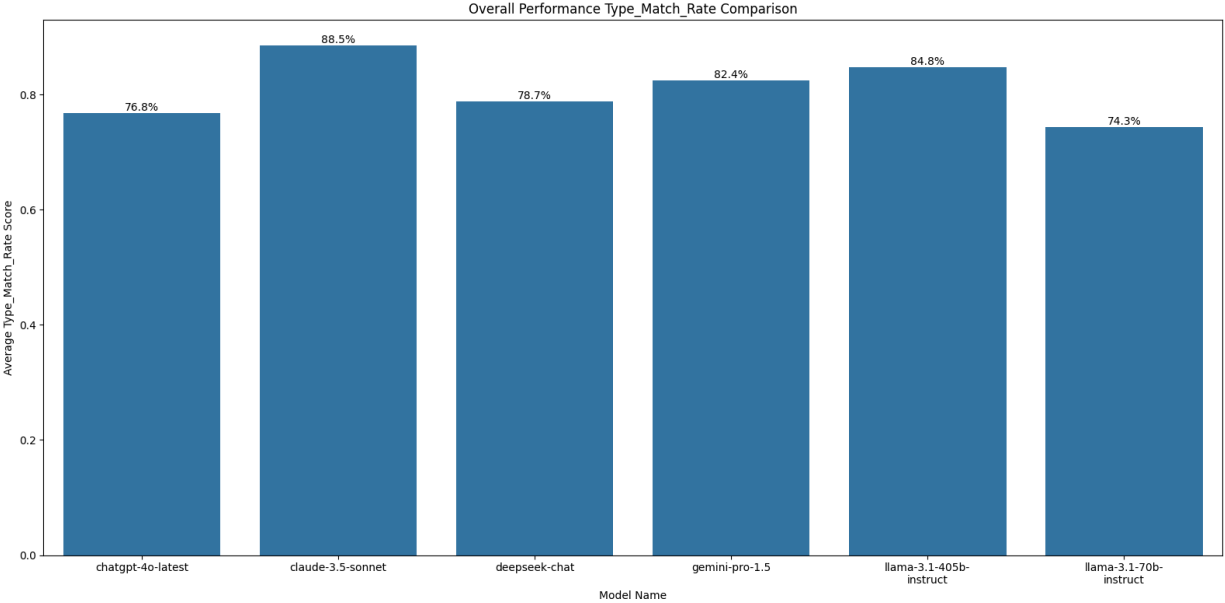

Figure 3: Type of source detection accuracy comparison.

However, the worth of accuracy on the type of source attribute is higher when the sourced statement accuracy is higher, as shown in Figure 4. Given Gemini 1.5 Pro significantly better score on sourced statement accuracy compared to Claude 3.5 Sonnet, if we combine the metrics, Gemini 1.5 Pro would be the better performer for a product of the two metrics, as a benchmark.

Measuring Large Language Models Capacity to Annotate Journalistic Sourcing

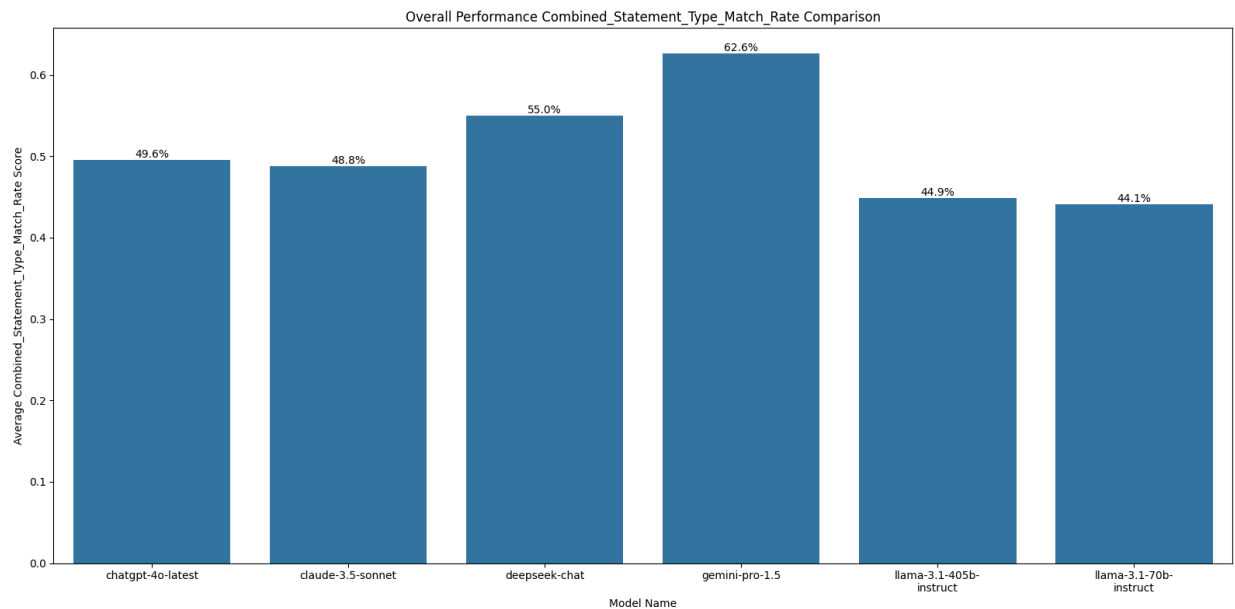

Figure 4: Comparing product of source statement and type accuracy.

**Name of source accuracy**

Name of source accuracy measures how effectively are the LLMs able to correctly pick up the name of the source by comparing with our ground truth data. As shown in Figure 5, we find that Claude 3.5 Sonnet, Gemini 1.5 Pro and Llama 3.1-405B are in the 80% range, DeepSeek R1 scored 75%, whereas GPT-4o and Llama 3.1-70B are roughly 72% accuracy.

Measuring Large Language Models Capacity to Annotate Journalistic Sourcing

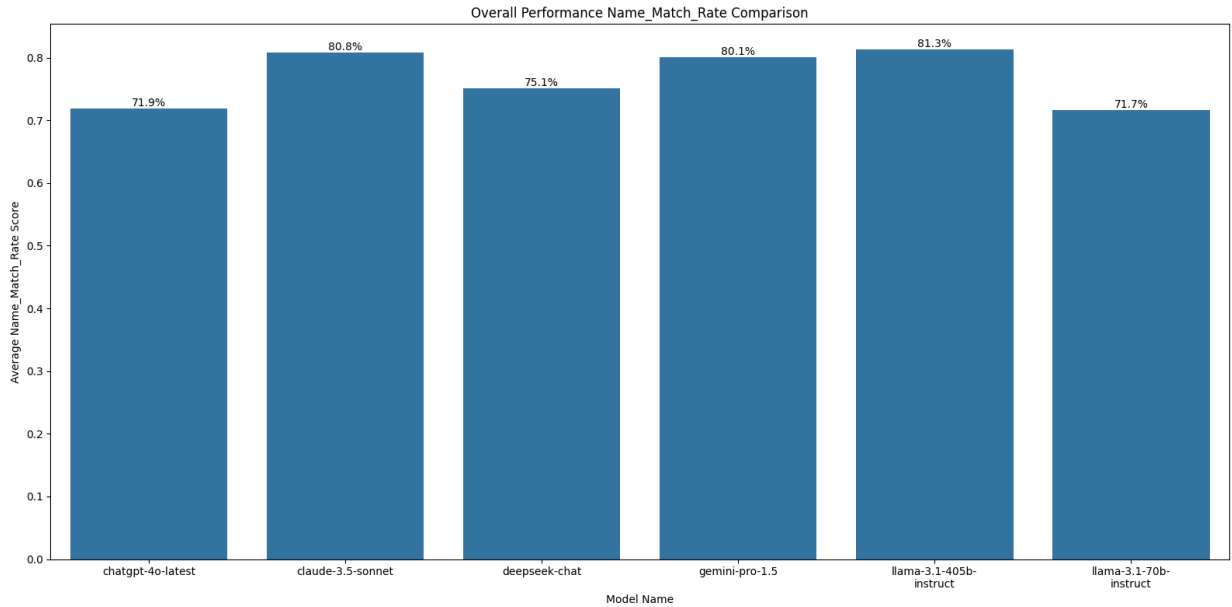

Figure 5: Name of source accuracy comparison.

These accuracy numbers go down marginally for all models if a condition is introduced. If the named source accuracy is calculated only for those cases where the type of source is also correct, Claude 3.5 Sonnet performs the best, at 78.6% accuracy, a little better than Gemini 1.5 Pro and Llama 3.1-405B. See Figure 6. The other three models score below 70%.

Measuring Large Language Models Capacity to Annotate Journalistic Sourcing

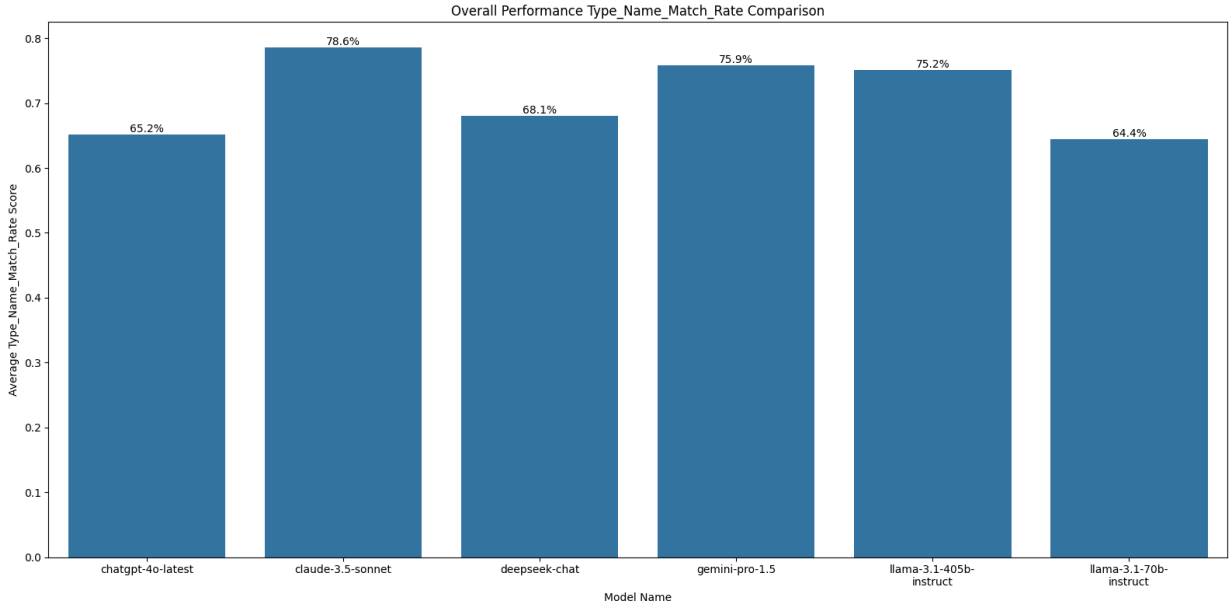

Figure 6: Name of source accuracy comparison for the condition that type of source is correct.

**Title of Source accuracy**

Title of source accuracy matches the titles of sources in the LLM generated data with the ground truth. As shown in Figure 7, the Llama 3.1-405B and Gemini 1.5 Pro score roughly 80% accuracy, with DeepSeek R1 and Claude 3.5 Sonnet in the 75-80% range.

Measuring Large Language Models Capacity to Annotate Journalistic Sourcing

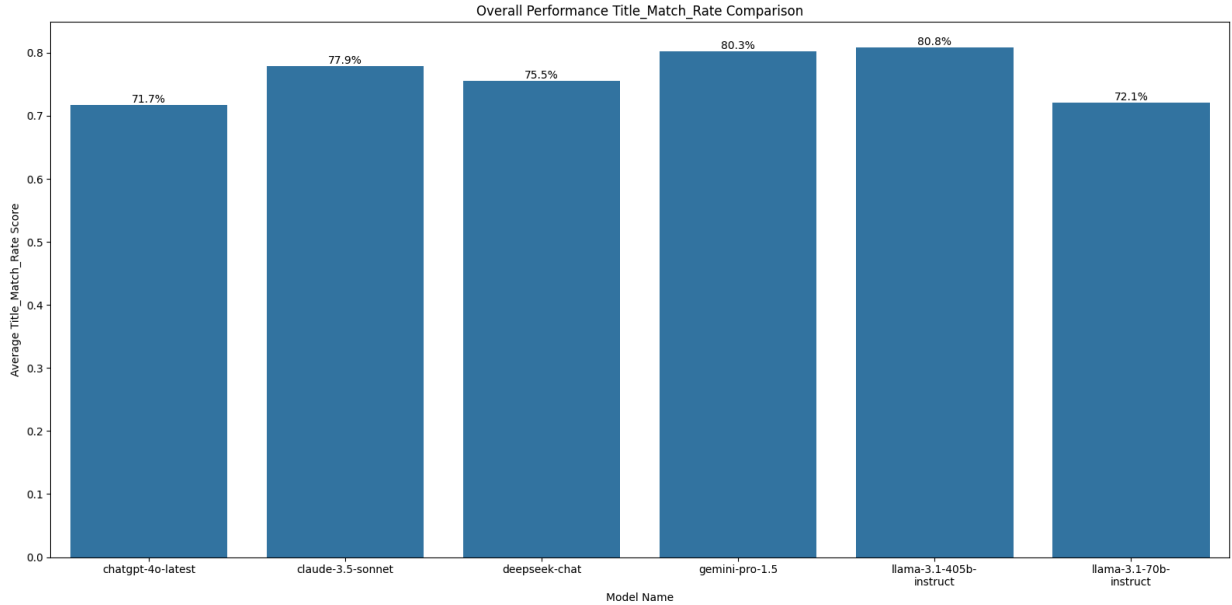

Figure 7: Source title accuracy comparison.

**Source Justification accuracy**

Source justification accuracy is scored by comparing source justification texts between the LLMs generated data and the ground truths. As shown in Figure 8, we find all the models here score poorly. Gemini 1.5 Pro and Claude 3.5 Sonnet reach 35% accuracy, while GPT-4o and DeepSeek R1 are in the 25-30% accuracy range. Llama 3.1-405B scores 23.7% and Llama 3.1-70B is very low.

Measuring Large Language Models Capacity to Annotate Journalistic Sourcing

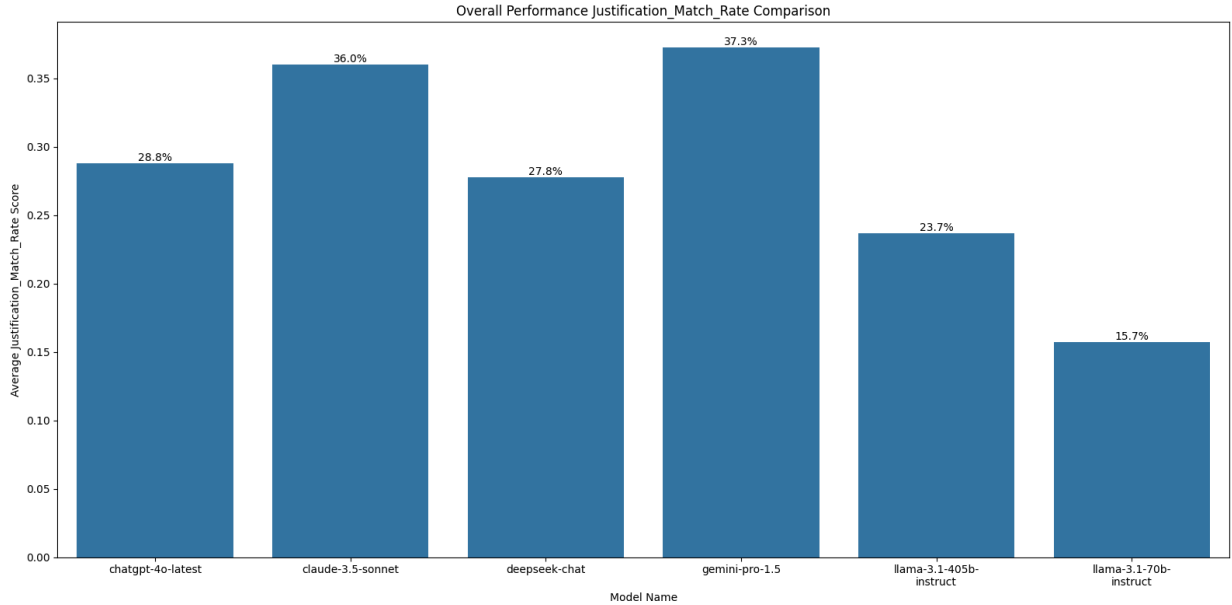

Figure 8: Source justification accuracy comparison.

One of the complexities here is how journalists use the title of a person. Sometimes the title of the source is offered as an inherent justification and this is evident in the sentence(s) used to introduce the source. We found that in some cases the LLMs were also adding the title to the justification column. Our intuition is that if the title and justification text data was combined and compared for similarity between the LLM data and ground truth, the accuracy score per mode for the merged justification attribute may change. Indeed it does, as shown in Figure 9. Claude 3.5 Sonnet received a roughly 50% accuracy score, with Gemini 1.5 Pro coming in after. However these are still low scores, indicating that LLMs are struggling with detecting unstructured journalistic signals in language using definitions.

Measuring Large Language Models Capacity to Annotate Journalistic Sourcing

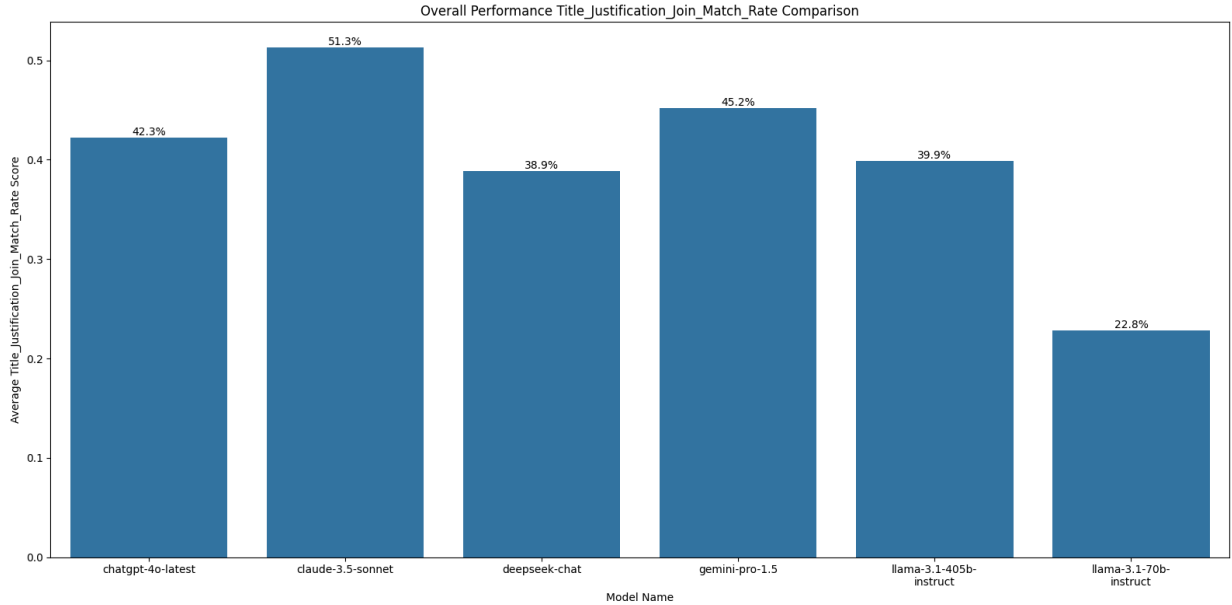

Figure 9. Revised source justification accuracy comparison, when title and justification are merged and compared with ground truth data.

**Overall model accuracy for journalistic sourcing**

We define overall model accuracy as the rate of LLMs getting all attributes (sourced statements, type of source, name of source, title of source and source justification) accurate for each story, as recorded in the ground truth files. As shown in Figure 10, we find that none of the models reach even 50%. Gemini Pro 1.5 scores significantly better here than DeepSeek R1, GPT-4o and Claude 3.5 Sonnet, whose scores are better than Llama 3.1-405B. Llama 3.1-70B has the poorest score.

Measuring Large Language Models Capacity to Annotate Journalistic Sourcing

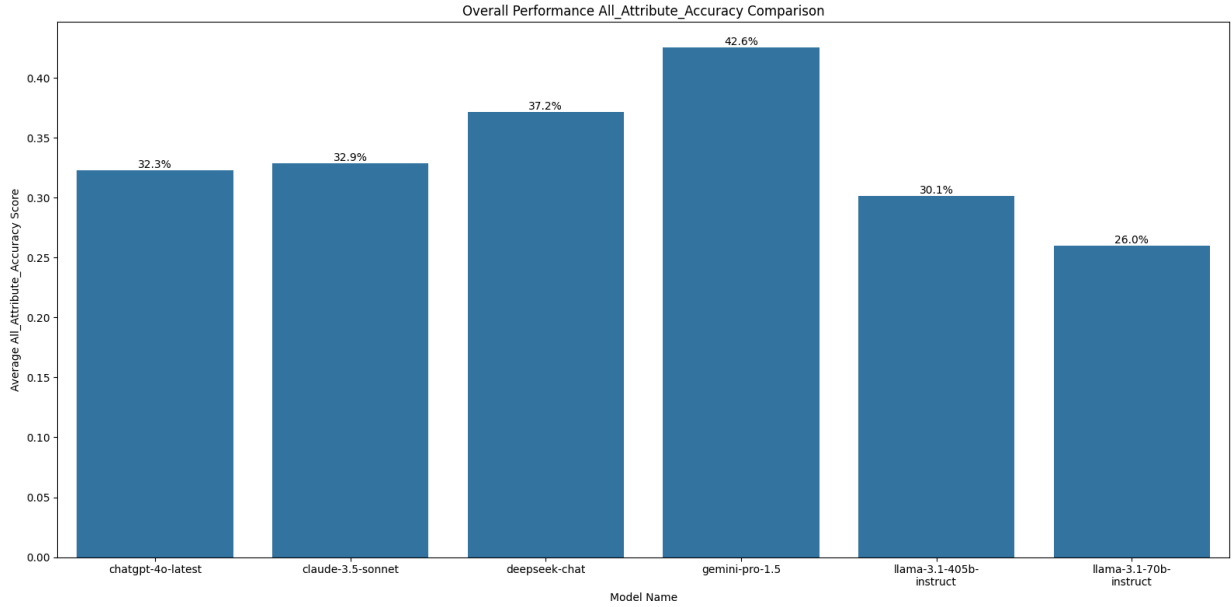

Figure 10: Comparison of overall accuracy models across all attributes.

|  | Statement_Match_Rate | Type_Match_Rate | Combined_Statement_Type_Match_Rate | Name_Match_Rate | Type_Name_Match_Rate | Title_Match_Rate | Justification_Match_Rate | Title_Justification_Join_Match_Rate | Overall accuracy |
|---|---|---|---|---|---|---|---|---|---|
| **model_name** | | | | | | | | | |
| **chatgpt-4o-latest** | 63.21% | 76.78% | 49.56% | 71.92% | 65.18% | 71.75% | 28.80% | 42.26% | 32.28% |
| **claude-3.5-sonnet** | 54.83% | **88.49%** | 48.78% | 80.83% | **78.58%** | 77.92% | 35.98% | **51.30%** | 32.90% |
| **deepseek-r1** | 69.38% | 78.73% | 54.97% | 75.08% | 68.06% | 75.49% | 27.76% | 38.86% | 37.19% |
| **gemini-pro-1.5** | **76.60%** | 82.40% | **62.63%** | 80.11% | 75.87% | 80.27% | **37.25%** | 45.22% | **42.55%** |
| **llama-3.1-405b-instruct** | 52.92% | 84.78% | 44.89% | **81.30%** | 75.18% | **80.84%** | 23.66% | 39.90% | 30.14% |
| **llama-3.1-70b-instruct** | 58.65% | 74.35% | 44.13% | 71.68% | 64.40% | 72.10% | **15.72%** | 22.78% | 25.98% |

Table 6: Combined table with all scores, across all sourcing attributes and overall score.

**Discussion**

Our key finding is that detecting sourced statements and types of sources is itself an area when LLMs have to make more progress than they are currently. The second key finding is that LLMs are currently unable to extract source justifications with high (say, 80%) accuracy, indicating room for much improvement ahead. Next, we review our findings with respect to each hypothesis we made at the outset.

**H1a**: LLMs will be able to accurately spot different types of statements attributed to sources, not only direct quotes.

**Finding:** While we do find that LLMs are able to go beyond direct quotes and detect a variety of different types of sourcing language and statements, only Gemini 1.5 Pro reaches even a modest 76.6% accuracy. The other models are in the 50-70% accuracy range.

**H1b**: (In conjunction with H1a) LLMs will be to identify the five different *types of sources* journalists use in the given a set of plain text definitions.

**Finding**: Overall, our finding is that LLMs are able to apply our definitions for the five types of sources. Claude 3.5 Sonnet, Gemini 1.5 Pro and Llama 3.1-405B all are in the 80-90% accuracy range, with Claude scoring 88.5%. and GPT-4o, DeepSeek R1, and Llama 3.1-70B are in the 70-75-80% accuracy range. One difficulty we see is with document sources that cite named people or named organizations within their content. We detailed these distinctions in our types of sources definitions given to the LLMs, but we did see evidence of conflation between document sources and named person sources for cases where the former is the correct annotation. This is already factored into the accuracy scores for the models.

**H2a**: LLMs will be able to identify source titles and justifications since this is a key test of the models' capacities on language.

**Finding**: We find that the five LLMs are able to detect source titles at high levels of accuracy (80%). In itself, this is not surprising because some NLP systems have already been able to do title detection (Full List of Annotators, n.d.). For source justifications, we find that there is a substantial distance to go in reaching high levels of accuracy. In some cases, we found that LLMs were synthesizing or summarizing justifications in their own words, instead of doing what the instructions asked them to: extract the exact words or sentences used by the journalist in the text. We had to add additional "do not" instructions in the user prompt for this.

**H2b**: LLMs will be able to spot anonymous sources as a type, related sourced statements, and extract the source justification as well.

**Finding**: Qualitatively, we find that LLMs are able to spot anonymous sourcing using the signals journalists put into the text about the source's role or presence or access to materials. The finding about this capacity has positive implications for journalism literacy. Anonymous sources are accepted by the public as a type of sourcing in journalism under particular contexts (Gottfried at al., 2024). News readers on the other hand feel that journalists are recklessly drawing claims and viewpoints from people whose identity is unknown. Digital news apps analyzing political stories could marshall this capacity carefully to signal the presence of anonymous sources and their justifications during news cycles on heated or sensitive political developments.

**H3:** We expect open-source models (Llama and DeepSeek R1 ), especially Llama's  405 Billion parameter model, to also perform as well as (or nearly) to closed source models (ChatGPT-4o, Claude 3.5 Sonnet, and Gemini 1.5 Pro).

Measuring Large Language Models Capacity to Annotate Journalistic Sourcing

**Finding:** Our central finding is that the open-source models Llama 3.1 -- 405B offers similar results to the ChatGPT-4o and Claude 3.5 Sonnet. Llama 3.1-70B scores poorest in the complex task of source justification annotations. Other studies that have compared open source LLMs to closed source suggest that fine-tuning may help (Alizadeh et al., 2024). But the distinctive overall (all-attributes) performance difference between Gemini Pro 1.5 and the other four models indicates that for the scenario of journalistic sourcing annotations, the type of model (open-source vs closed) is not a significant driver of accuracy.

## Limitations and future work

1. In our current framework, we intentionally limited the range in types of sources to five. In reality, documents as a type of source are amenable to a deeper taxonomy. For instance, it is now common practice in digital journalism to use inline text to link directly to primary documents online (URLs of documents issued by official authorities). The document may not necessarily be suggestively referred to as a verdict (courts), report, study, research paper, etc. But at the same time, journalists also use inline text to add secondary sourcing URLs to other press articles, often to the same news outlet's earlier coverage. There is also the case of citations to other web pages that fit the definition of a document source. Our current work only includes document sources referred to in plain text. We plan to disambiguate and expand document sourcing to a deeper taxonomy that includes sourcing signals through URLs, for a future effort.

2. Our proportions of different types of sourcing - in the 34 article sample set - was not equally distributed across all five types. Named people and named organizations are a very high proportion in our source types. Future experiments would need higher proportions of anonymous and document sourcing in the samples to develop sub-benchmarks for those types of sourcing.

3. Unnamed individuals (without the reporter qualifying in the text as having sought and received anonymity) as a general type of source: A non-standardized practice in journalistic sourcing is to refer to an individual as a source, not name them, but also not disclose a specific justification (or qualification) for anonymity. This type of attribution is different from the references to "unnamed groups of people" as a source when journalists witness an event or development and report a group of people saying something. Defining a separate type of source called "unnamed individual" for this case will have the advantage of testing whether a) LLMs can distinguish between all three types of sources and b) whether this can then be used to track and callout instances in news cycles where stories are carrying claims and statements from unnamed individuals without specific conditions of anonymity being transparently disclosed in the story.

4. LLMs temperature setting: We set the temperature to zero for all LLMs we evaluated. It is possible that by increasing the temperature setting marginally from 0 to 0.1 or 0.2, the LLMs may produce variations in results for one or more of the sourcing attributes. While most model literature recommends zero for temperature settings that map to deterministic behavior, GPT-4o has recommended the low-end to be 0.2 and not 0. In future work, we plan to examine whether LLMs will show improvement (or regress) for accuracy metrics in sourced statements and source justification annotations, with revisions to the temperature setting.

**Conclusion**

One area that has not received sufficient scenario development attention for LLM capacity benchmarking is real-world journalism, and in particular journalistic sourcing and ethics. Evaluating the capacities of LLMs to annotate the sourcing in stories is the core work

Measuring Large Language Models Capacity to Annotate Journalistic Sourcing

behind this paper. We claim that our work adds a new scenario to benchmark LLMs for journalism which we draw from the real world of everyday journalistic sourcing work. We believe that this is a scenario that warrants a benchmark approach because this will help evaluate the various models along this new dimension. We offer the use case, our dataset[1] of articles, our ground truth data, the LLM prompts, LLM generated data, and a set comparison metrics for each model as the first step towards systematic benchmarking. The long-term implications are that if some LLMs could get the job done, digital news applications may be able to perform more complex journalistic source evaluations during news cycles in real time.

**Ethics Statement**

Our study is about language models and their benchmarking for journalistic sourcing annotation. Our use of text from the selected news articles and the sourcing statements is strictly for their language patterns and data for our experiment. We have not used the content for any other publication or evaluation purposes. We are also not making any study or claims or observations about the ethics of sourcing in these individual news articles or of their authoring journalists themselves and hence we did not engage in actively disclosing our analysis to each news outlet whose stories we included for this experiment.

**Disclosure on Conflicts of Interest**

There are no relevant financial or non-financial competing interests to report for this study.

**Acknowledgments**

Djordje Padeski provided helpful comments on the first version.

Measuring Large Language Models Capacity to Annotate Journalistic Sourcing

# References

Bommasani, R., Liang, P., & Lee, T. (2023). Holistic evaluation of language models. *Annals of the New York Academy of Sciences*, *1525*(1), 140–146. https://doi.org/10.1111/nyas.15007

Steensen, S., Belair-Gagnon, V., Graves, L., Kalsnes, B., & Westlund, O. (2022). Journalism and source criticism. Revised Approaches to Assessing Truth-Claims. *Journalism Studies*, *23*(16), 2119–2137. https://doi.org/10.1080/1461670x.2022.2140446

Vincent, S. (2023). Reorienting journalism to favor democratic agency. In *Reinventing Journalism to Strengthen Democracy: Insights from Innovators*. (pp. 21–50). Kettering Foundation.

*Holistic Evaluation of Language Models (HELM)*. (n.d.). https://crfm.stanford.edu/helm/classic/latest/#/scenarios

Moorhead, L. (2024). Who Gets a Voice in Homelessness? A Content Analysis of Quotations Used by Journalists. *Journalism Practice*, 1–28. https://doi.org/10.1080/17512786.2023.2300281

Jaakkola, M. (2018). Journalistic writing and style. *Oxford Research Encyclopedia of Communication*. https://doi.org/10.1093/acrefore/9780190228613.013.884

Gans, H. J. (2004). *Deciding what's news*. Northwestern University Press.

Bhargava, R. &. H. E. &. H. M. (2024). Testing generative AI for source audits in Student-Produced Local news. *ideas.repec.org*. https://ideas.repec.org/p/osf/socarx/7hc2d.html

Li, C., Diakopoulos, N., Northwestern University, & Northwestern University. (2024). Probing GPT-4 for knowledge of Journalistic Tasks. *Proceedings of Computation + Journalism Symposium (C+J Symposium)*, 5. https://cplusj2024.github.io/papers/CJ_2024_paper_17.pdf

Measuring Large Language Models Capacity to Annotate Journalistic Sourcing

Zhang, T., Ladhak, F., Durmus, E., Liang, P., McKeown, K., & Hashimoto, T. B. (2024). Benchmarking large language models for news summarization. Transactions of the Association for Computational Linguistics, 12, 39–57. https://doi.org/10.1162/tacl_a_00632

Spangher, A., Youn, J., DeButts, M., Peng, N., Ferrara, E., & May, J. (2024, November 7). *Explaining mixtures of sources in news articles*. arXiv.org. https://arxiv.org/abs/2411.05192

Spangher, A., Peng, N., Ferrara, E., & May, J. (2023, December). Identifying Informational Sources in News Articles. In *Proceedings of the 2023 Conference on Empirical Methods in Natural Language Processing* (pp. 3626-3639).

Muzny, G., Fang, M., Chang, A., & Jurafsky, D. (2021). A Two-stage Sieve Approach for Quote Attribution. *ACL Anthology*. https://aclanthology.org/E17-1044/

Vincent, S., Wu, X., Huang, M., & Fang, Yi. (2023). Could Quoting Data Patterns Help in Identifying Journalistic Behavior Online? *In Official Research Journal of the International Symposium on Online Journalism* (Vol. 13, Issue 1, p. 9) [Journal-article].

Shang, X., Peng, Z., Yuan, Q., Khan, S., Xie, L., Fang, Y., & Vincent, S. (2022). DIANES. *Proceedings of the 45th International ACM SIGIR Conference on Research and Development in Information Retrieval*. https://doi.org/10.1145/3477495.3531660

Wang, J., (2024). A Step Towards Automated Ethical Analysis in Journalism: Measuring LLMs' Performance in Extracting Sourcing Information" (2024). Computer Science and Engineering Master's Theses. 43. https://scholarcommons.scu.edu/cseng_mstr/43

*Full list of annotators*. (n.d.). CoreNLP. https://stanfordnlp.github.io/CoreNLP/annotators.html

Gottfried, J., Walker, M., (2024, April 14). Most Americans see a place for anonymous sources in news stories, but not all the time. *Pew Research Center*.

Measuring Large Language Models Capacity to Annotate Journalistic Sourcing

https://www.pewresearch.org/short-reads/2020/10/09/most-americans-see-a-place-for-anonymous-sources-in-news-stories-but-not-all-the-time/

Brigham, N. G., Gao, C., Kohno, T., Roesner, F., & Mireshghallah, N. (2024). Breaking News: Case Studies of Generative AI's Use in Journalism. ArXiv.org. https://arxiv.org/abs/2406.13706

Guha, A. (2024, January 31). Vermont House overwhelmingly backs bill prohibiting race-based hair discrimination. VTDigger. https://vtdigger.org/2024/01/31/vermont-house-overwhelmingly-backs-bill-prohibiting-race-based-hair-discrimination/

Wei, J., Wang, X., Schuurmans, D., Bosma, M., Ichter, B., Xia, F., Chi, E., Le, Q., & Zhou, D. (2022). Chain of Thought Prompting Elicits Reasoning in Large Language Models. ArXiv:2201.11903 [Cs]. https://arxiv.org/abs/2201.11903

Mouselimis L (2021). *fuzzywuzzyR: Fuzzy String Matching*. R package version 1.0.5, https://CRAN.R-project.org/package=fuzzywuzzyR.

Reimers, N., & Gurevych, I. (2019). Sentence-BERT: Sentence Embeddings using Siamese BERT-Networks. Conference on Empirical Methods in Natural Language Processing.

Duffy, M. J. (2014). Anonymous Sources: A historical review of the norms surrounding their use. *American Journalism*, *31*(2), 236–261. https://doi.org/10.1080/08821127.2014.905363

Carlson, M. (2010). Whither anonymity? Journalism and unnamed sources in a changing media environment. In *Journalists, Sources, and Credibility* (1st Edition, pp. 49–60). https://doi.org/10.4324/9780203835708-9

Alizadeh, M., Kubli, M., Samei, Z., Dehghani, S., Zahedivafa, M., Bermeo, J. D., Korobeynikova, M., & Gilardi, F. (2024). Open-source LLMs for text annotation: a practical

Measuring Large Language Models Capacity to Annotate Journalistic Sourcing

guide for model setting and fine-tuning. *Journal of Computational Social Science*, *8*(1). https://doi.org/10.1007/s42001-024-00345-9

A. Zhukova, T. Ruas, F. Hamborg, K. Donnay and B. Gipp, "What's in the News? Towards Identification of Bias by Commission, Omission, and Source Selection (COSS)," *2023 ACM/IEEE Joint Conference on Digital Libraries (JCDL)*, Santa Fe, NM, USA, 2023, pp. 258-259, doi: 10.1109/JCDL57899.2023.00050.

## List of Tables

## List of Figures

Figure 9. Revised source justification accuracy comparison, when title and justification are merged and compared with ground truth data.

Figure 10: Comparison of overall accuracy models across all attributes.